\documentclass{article}
\usepackage{spconf}
\usepackage{amsmath,amsfonts,amssymb,amsthm}
\usepackage{bm}
\usepackage{mathtools}
\usepackage{graphicx}
\usepackage{url}
\usepackage{enumitem}
\usepackage{color}
\usepackage{booktabs}
\usepackage{multirow}

\usepackage{algorithm}
\usepackage{algorithmic}

\usepackage[font=small,skip=3pt]{caption}

\def\0{{\mathbf 0}}
\def\1{{\mathbf 1}}

\def\a{{\mathbf a}}

\def\e{{\mathbf e}}
\def\f{{\mathbf f}}

\def\s{{\mathbf s}}

\def\x{{\mathbf x}}
\def\y{{\mathbf y}}

\def\B{{\mathbf B}}

\def\D{{\mathbf D}}

\def\H{{\mathbf H}}

\def\LP{{\mathbf L}}
\def\M{{\mathbf M}}

\def\S{{\mathbf S}}

\def\W{{\mathbf W}}

\def\ie{{\textit{i.e.}}}

\def\cA{{\mathcal A}}

\def\cE{{\mathcal E}}

\def\cG{{\mathcal G}}

\def\cK{{\mathcal K}}
\def\cL{{\mathcal L}}

\def\cO{{\mathcal O}}

\def\cU{{\mathcal U}}
\def\cV{{\mathcal V}}

\theoremstyle{plain}

\newtheorem{lemma}{Lemma}

\theoremstyle{definition}

\theoremstyle{definition}

\newlength{\bibitemsep}
\newlength{\bibparskip}
\let\oldthebibliography\thebibliography
\renewcommand\thebibliography[1]{\oldthebibliography{#1}\setlength{\parskip}{\bibitemsep}\setlength{\itemsep}{\bibparskip}}

\title{Fast Graph Sampling for Short Video Summarization using \\ Gershgorin Disc Alignment}
\name{Sadid Sahami$^{\dagger}$
\qquad Gene Cheung$^{\star}$ 
\qquad Chia-Wen Lin$^{\dagger}$ 
\thanks{Gene Cheung acknowledges the support of the NSERC grants RGPIN-2019-06271,  RGPAS-2019-00110.}
}

\address{$^{\star}$ York University, Toronto, Canada ~~~~~~~~~~
$^{\dagger}$ National Tsing Hua University, Hsinchu, Taiwan}

\begin{document}

\ninept

\maketitle
\begin{abstract}
We study the problem of efficiently summarizing a short video into several keyframes, 
leveraging recent progress in fast graph sampling. 
Specifically, we first construct a similarity path graph (SPG) $\cG$, represented by graph Laplacian matrix $\LP$, where the similarities between adjacent frames are encoded as positive edge weights.
We show that maximizing the smallest eigenvalue $\lambda_{\min}(\B)$ of a coefficient matrix $\B = \text{diag}(\a) + \mu \LP$, where $\a$ is the binary keyframe selection vector, is equivalent to minimizing a worst-case signal reconstruction error.
We prove that, after partitioning $\cG$ into $Q$ sub-graphs $\{\cG^q\}^Q_{q=1}$, the smallest Gershgorin circle theorem (GCT) lower bound of $Q$ corresponding coefficient matrices---$\min_q \lambda^-_{\min}(\B^q)$ ---is a lower bound for $\lambda_{\min}(\B)$.
This inspires a fast graph sampling algorithm to iteratively partition $\cG$ into $Q$ sub-graphs using $Q$ samples (keyframes), while maximizing $\lambda^-_{\min}(\B^q)$ for each sub-graph $\cG^q$.
Experimental results show that our algorithm achieves comparable video summarization performance as state-of-the-art methods, at a substantially reduced complexity.
\end{abstract}


\vspace{-0.05in}
\section{INTRODUCTION}
\label{sec:intro}
The ubiquity of mobile phones equipped with high-resolution, high-quality cameras means that video contents (typically of short duration no longer than a handful of minutes) are now generated at an astonishing rate\footnote{As of Feb 2020, 500 hours of videos were uploaded to YouTube every minute \url{https://blog.youtube/news-and-events/youtube-at-15-my-personal-journey/}.}. On the other hand, human's available resources to consume media content, both in time and attention, are far more limited, and thus it is imperative to automatically summarize uploaded videos quickly into several keyframes to facilitate downstream tasks such as a user's selection, retrieval~\cite{hu2011survey}, and classification~\cite{brezeale2008automatic}. 

Keyframe selection for short videos is well studied in the literature \cite{yihonggong2000video,luo2009extracting}.
One main category of algorithms is clustering-based \cite{mundur2006keyframebased,furini2009stimo,deavila2011vsumm,anirudh2016diversity,wu2017novel}: grouping frames into clusters, then selecting a representative frame from each cluster.
While some methods have low complexity~\cite{deavila2011vsumm,furini2009stimo}, they are heuristic-based when setting the number of clusters and selecting keyframes.
In contrast, there exist optimization-based methods, such as recent \textit{sparse dictionary selection} methods~\cite{mei2021patcha,mei2014l2,mei2015video}.
These methods formulated video summarization as a sparse dictionary learning problem by selecting a subset of frames (keyframes) as the dictionary for minimum-error reconstruction.
The sparse dictionary selection problem is, however, NP-hard \cite{tillmann2015computational}.
To solve the problem, these methods resorted to an approximate solution by selecting dictionary atoms (keyframes) iteratively based on a pre-defined criterion.
This also necessitated an initialization by selecting the first keyframe (\text{e.g.}, the first frame).
A recent state-of-the-art method \cite{mei2021patcha} extended the formulation to model blocks of image patches per frame (SBOMP) and further grouped neighboring frame blocks into super-blocks (SBOMPn), where the atoms for the dictionary are greedily selected by a variation of \textit{Orthogonal Matching Pursuit} (OMP).
This complex spatio-temporal representation resulted in better performance at a cost of increased computational complexity.

Keyframe selection bears a strong resemblance to the \textit{graph sampling} problem in \textit{graph signal processing} (GSP) \cite{ortega18ieee,cheung18}: 
given a combinatorial graph with edge weights that reflect similarities of connected node pairs, choosing a node subset to collect signal samples, so that the reconstruction quality of an assumed smooth (or low-pass) graph signal is optimized.
In particular, a recent fast graph sampling scheme called \textit{Gershgorin disc alignment sampling} (GDAS)~\cite{bai2020fast}, based on the well-known \textit{Gershgorin circle theorem} (GCT)~\cite{varga04}, achieved competitive performance while running in roughly linear time. 
Leveraging GDAS, we propose an efficient keyframe extraction algorithm, \textit{the first in the literature to pose video summarization as a graph sampling problem}.

Specifically, we first construct a \textit{similarity path graph} (SPG) $\cG$ for frames in a video, where each positive edge weight $w_{i,i+1}$ is inversely proportional to \textit{feature distance} between feature vectors $\f_i$ and $\f_{i+1}$ (computed using GoogLeNet features~\cite{szegedy2015going} pre-trained on ImageNet~\cite{deng2009imagenet}) associated with frames $i$ and $i+1$, respectively.
Given graph Laplacian matrix $\LP$ corresponding to SPG $\cG$, we show that maximizing the smallest eigenvalue $\lambda_{\min}(\B)$ of a coefficient matrix $\B = \text{diag}(\a) + \mu \LP$, where $\a$ is the binary keyframe selection vector, is equivalent to minimizing a worst-case signal reconstruction error.
Next, we prove that, after partitioning $\cG$ into $Q$ sub-graphs $\{\cG^q\}_{q=1}^Q$, the smallest GCT lower bound of $Q$ corresponding coefficient matrices---$\min_q \lambda^-_{\min}(\B^q)$ ---is a lower bound for $\lambda_{\min}(\B)$.
This inspires a fast graph sampling algorithm similar to GDAS \cite{bai2020fast} to partition $\cG$ into $Q$ sub-graphs, while maximizing $\lambda^-_{\min}(\B^q)$ for each sub-graph $\cG^q$.
Experimental results show that our algorithm achieves comparable video summarization performance as state-of-the-art methods~\cite{mei2014l2,cong2017adaptive,mei2021patcha}, at a substantially reduced complexity.

\vspace{-0.11in}
\section{PRELIMINARIES}
\label{sec:prelim}
\vspace{-0.05in}
\subsection{Graph Definitions}

A positive graph $\cG(\cV,\cE,\W)$ is defined by a set of $N$ nodes $\cV = \{1, \ldots, N\}$, edges $\cE = \{(i,j)\}$, and an \textit{adjacency matrix} $\W$, where $W_{i,j} = w_{i,j} \in \mathbb{R}^+$ is the positive edge weight if $(i,j) \in \cE$, and $W_{i,j} = 0$ otherwise. 
If self-loops exist, then $W_{i,i} \in \mathbb{R}^+$ is the self-loop weight for node $i$.
\textit{Degree matrix} $\D$ has diagonal entries $D_{i,i} = \sum_{j} W_{i,j}, \forall i$. 
A \textit{combinatorial graph Laplacian matrix} $\LP$ is defined as $\LP \triangleq \D - \W$ \cite{ortega18ieee}. 
One can easily show that $\LP$ is \textit{positive semi-definite} (PSD), \ie, all eigenvalues of $\LP$ are non-negative \cite{cheung18}. 
If self-loops exist, then the \textit{generalized graph Laplacian matrix} $\cL$ is $\cL \triangleq \D - \W + \text{diag}(\W)$ to account for self-loops.

A signal $\x \in \mathbb{R}^n$ is smooth w.r.t. graph $\cG$ (without self-loops) if the \textit{graph Laplacian regularizer} (GLR), $\x^{\top} \LP \x$, is small \cite{pang17}: 
\begin{align}
\x^{\top} \LP \x = \sum_{(i,j) \in \cE} W_{i,j} (x_i - x_j)^2 .
\end{align}
GLR is used to regularize ill-posed inverse problems such as graph signal denoising, dequantization, and interpolation \cite{pang17,liu17,chen21}.

\subsection{Gershgorin Circle Theorem}
\label{subsec:GCT}

Given a real symmetric matrix $\M$, corresponding to each row $i$ is a \textit{Gershgorin disc} $i$ with center $c_i \triangleq M_{i,i}$ and radius $r_i \triangleq \sum_{j \neq i} |M_{i,j}|$. 
GCT \cite{varga04} states that each eigenvalue $\lambda$ resides in at least one Gershgorin disc. A corollary is that the smallest Gershgorin disc left-end $\lambda^-_{\min}(\M)$ is a lower bound of the smallest eigenvalue $\lambda_{\min}(\M)$ of $\M$, \ie,
\begin{align}
\lambda^-_{\min}(\M) \triangleq \min_i c_i - r_i \leq \lambda_{\min}(\M) .
\label{eq:GCT}
\end{align}
Note that a \textit{similarity transform} \cite{horn2012matrix} $\S \M \S^{-1}$ for an invertible matrix $\S$ has the same set of eigenvalues as $\M$.
Thus, a GCT lower bound for $\S \M \S^{-1}$ is also a lower bound for $\M$, \ie, 
\begin{align}
\lambda^-_{\min}(\S \M \S^{-1}) \leq \lambda_{\min}(\M) .
\label{eq:GCT2}
\end{align}
Because GCT lower bound of the original matrix $\M$ \eqref{eq:GCT} is often loose, typically a suitable similarity transform is first performed to obtain a tighter bound \eqref{eq:GCT2}. 
 
\vspace{-0.11in}
\section{PATH GRAPH CONSTRUCTION}
\label{sec:graph}

To represent video frames, we utilize GoogLeNet, pretrained on ImageNet~\cite{deng2009imagenet}.
Each frame $F_i$ of a video sequence is fed into the pre-trained network, and the activations of $\textrm{pool5}$ layer in the network are recorded as a \textit{feature vector} $\f_i \in \mathbb{R}^K$.

Given extracted feature vectors $\f_i$ and $\f_{i+1}$ for frames $F_i$ and $F_{i+1}$, we compute edge weight $w_{i,i+1}$ of a \textit{similarity path graph} (SPG) $\cG$ connecting nodes $i$ and $i+1$ as follows.
We first compute \textit{feature distance} $\delta_{i,i+1}$ between $\f_i$ and $\f_{i+1}$ as

\vspace{-0.05in}
\begin{small}
\begin{align}
\delta_{i,i+1} = \frac{\|\f_i - \cos(\theta_{i,i+1})\f_{i+1}\|_2 + \|\f_{i+1} - \cos(\theta_{i,i+1})\f_i\|_2}{\|\f_i\|_2 + \|\f_{i+1}\|_2}
\end{align}
\end{small}\noindent
where $\theta_{i,i+1}$ is the angle between $\f_i$ and $\f_{i+1}$.
The distance definition $\delta_{i,i+1}$ is symmetric: \ie, $\delta_{i,i+1} = \delta_{i+1,i}$.
Note that if $\f_i = \f_{i+1}$, \ie, frames $i$ and $i+1$ are identical, then $\cos (\theta_{i,i+1}) = 1$, and $\delta_{i,i+1} = 0$.
We define the edge weight connecting frames $i$ and $i+1$ as $w_{i,i+1} = 1-\delta_{i,i+1}$.
Since the number of edges in an $N$-node SPG $\cG$ is $N-1$, the complexity of graph construction is $\cO(N)$.

\vspace{-0.11in}
\section{PATH GRAPH SAMPLING}
\label{sec:sampling}
\subsection{Defining Sampling Objective}

Given SPG $\cG$, we next derive an objective to choose sample nodes in $\cG$. 
Denote by $\LP \in \mathbb{R}^{N \times N}$ a combinatorial graph Laplacian matrix corresponding to $\cG$, and by $\H \in \{0,1\}^{C \times N}$ a \textit{sampling matrix}--- each row being a one-hot vector---that chooses $C$ samples from a signal $\x \in \mathbb{R}^N$.
As an example, the following sampling matrix $\H$ chooses nodes $1$ and $3$ from a four-node graph $\cG$:
\begin{align}
\H = \left[ \begin{array}{cccc}
1 & 0 & 0 & 0 \\
0 & 0 & 1 & 0
\end{array} \right] .
\label{eq:exH}
\end{align}
Denote by $\y \in \mathbb{R}^C$ a vector of $C$ observed samples.
Using GLR for regularization \cite{pang17}, signal $\x$ can be estimated by solving the following optimization:
\begin{align}
\min_{\x} \|\y - \H \x \|^2_2 + \mu \x^{\top} \LP \x
\label{eq:reconObj}
\end{align}
where $\mu > 0$ is a weight parameter.

Since \eqref{eq:reconObj} is quadratic, convex and differentiable, the solution $\x^*$ can be obtained by solving the following linear system:
\begin{align}
\left( \H^{\top} \H + \mu \LP \right) \x^* = \H^{\top} \y
\end{align}
where the \textit{coefficient matrix} $\B = \H^{\top} \H + \mu \LP$ is \textit{positive definite} (PD) and thus invertible \cite{bai2020fast}. 
$\H^{\top} \H$ is a diagonal matrix with the diagonal entries corresponding to chosen samples equal to $1$.
Thus, one can compactly rewrite $\H^{\top} \H = \text{diag}(\a)$, where $\a \in \{0,1\}^N$ is a \textit{sample selection} vector with those entries corresponding to chosen samples equal to $1$.

One can show that maximizing the smallest eigenvalue $\lambda_{\min}(\B)$ of coefficient matrix $\B$---known as the \textit{E-optimality criterion} in system design \cite{ehrenfeld1955efficiency}---minimizes the worst-case reconstruction error (Proposition 1 in \cite{bai2020fast}). 
For ease of computation, we maximize instead its GCT lower bound $\lambda^-_{\min}(\S \B \S^{-1})$:
\begin{align}
\max_{\a, \S} \lambda^-_{\min} \left( \S (\text{diag}(\a) + \mu \LP)  \S^{-1} \right), 
~~\mbox{s.t.}~\|\a\|_1 \leq C 
\label{eq:samplingPrimal}
\end{align}
where $\a \in \{0, 1\}^N$ and $\S$ is any invertible matrix. 
For simplicity, we focus on diagonal matrices, \ie, $\S = \text{diag}(\s)$, where $s_i > 0, \forall i$.

\subsection{Problem Transformation}

Since \eqref{eq:samplingPrimal} is NP-hard \cite{bai2020fast}, for expediency, we swap the roles of the objective and constraint in \eqref{eq:samplingPrimal} and solve instead its corresponding \textit{dual problem} given threshold $0 < T <1$ \cite{bai2020fast}, \ie,
\begin{align}
\min_{\a, \S} \|\a\|_1,
~~\mbox{s.t.}~~ \lambda^-_{\min} \left( \S (\text{diag}(\a) + \mu \LP) \S^{-1} \right) \geq T .
\label{eq:samplingDual}
\end{align}
where $T$ is inversely proportional to $\|\a^*\|_1$ of the optimal solution $\a^*$ to \eqref{eq:samplingDual} \cite{bai2020fast}. 
Thus, to find the dual optimal solution $\a^*$ such that $\|\a^*\|_1 = C$---and hence the optimal solution to the primal problem \eqref{eq:samplingPrimal} also---one can perform binary search on $T$ to find an appropriate $T$. 
We design a fast algorithm to approximately solve \eqref{eq:samplingDual} for a given $T$ next.

We first state the following motivating lemma.
Given a positive graph $\cG$ (possibly with self-loops), suppose we partition it into $Q$ sub-graphs $\cG^1, \ldots, \cG^Q$, where edges connecting nodes from different sub-graphs are removed.

\begin{lemma}
The smallest GCT lower bound among all generalized graph Laplacians $\cL^1, \ldots, \cL^Q$ for sub-graphs $\cG^1, \ldots, \cG^Q$ of graph $\cG$ is a lower bound for $\lambda_{\min}(\cL)$ of $\cG$, \ie,
\begin{align}
\min_{q \in \{1, \ldots, Q\}} \lambda^-_{\min}(\cL^q) \leq \lambda_{\min} (\cL) .
\label{eq:distGCT_lower}
\end{align}
\label{lemma:lemma1}
\end{lemma}

\vspace{-0.2in}
\begin{proof}
We prove \eqref{eq:distGCT_lower} as follows.
By GCT, we know that $\lambda^-_{\min}(\cL) \leq \lambda_{\min}(\cL)$.
Left-end of disc $i$ of $\cL$ is the disc center minus the radius, \ie,
\begin{align}
c_i(\cL) - r_i(\cL) &= \cL_{i,i} - \sum_{j \neq i} |\cL_{i,j}| \\
&= W_{i,i} + \sum_{j \neq i} W_{i,j} - \sum_{j \neq i} W_{i,j} = W_{i,i} .
\nonumber
\end{align}
Suppose now node $i$ is in sub-graph $\cG^q$.
Suppose further that one or more edges $(i,k)$, $k \in \cK$, are removed during sub-graph partition.
The left-end of row $i$ in generalized graph Laplacian matrix $\cL^q$ of $\cG^q$ is
\begin{align}
c_i(\cL^q) - r_i(\cL^q) &= \cL^q_{i,i} - \sum_{j \neq i} |\cL^q_{i,j}| \\
&= W_{i,i} + \!\!\!\! \sum_{j \not\in \cK \cup \{i\}} \!\! W_{i,j} -  \!\!\!\! \sum_{j \not\in \cK \cup \{i\}} \!\! W_{i,j} = W_{i,i} .
\nonumber
\end{align}
We thus conclude that the left-ends of row $i$ in $\cL$ and $\cL^q$ are the same.
Since $\min_q \lambda^-_{\min}(\cL^q) = \min_q \min_i c_i(\cL^q) - r_i(\cL^q)$ is the same as $\min_i c_i(\cL) - r_i(\cL) = \lambda^-_{\min}(\cL)$, we conclude \eqref{eq:distGCT_lower} is true.
\end{proof}
The corollary of the lemma is that one can maximize a lower bound of $\lambda_{\min}(\cL)$ by maximizing individual GCT lower bounds $\lambda^-_{\min}(\cL^q)$ of sub-graph Laplacians $\cL^q$'s.

\subsection{Line Graph Sampling Algorithm}

Given Lemma\;\ref{lemma:lemma1}, we solve \eqref{eq:samplingDual} by dividing graph $\cG$ into sub-graphs, where \textit{each} sub-graph $\cG^q$, bestowed with a single node sample, satisfies $\lambda^-_{\min}(\S \B^q \S^{-1}) \geq T$ for some $\S$. 
Specifically, given SPG $\cG$ with $N$ nodes, we first identify sample $k$ solving the following optimization:
\begin{align}
\max_{k, \S}~ d, ~~
\mbox{s.t.} ~~\lambda^-_{\min}(\S (\text{diag}(\e_k) + \mu \LP^d) \S^{-1}) \geq T
\label{eq:sampleOpt}
\end{align}
where $\e_k$ is a \textit{canonical vector} of zeros except entry $k$ which equals $1$, and $\LP^{d}$ is the graph Laplacian matrix corresponding to the sub-graph of the first $d$ nodes.
The idea is to find a sample node $k$ which can ``cover" the \textit{longest} sub-graph from $1$ to $d$, \ie, satisfying $\lambda^-_{\min}(\S \B \S^{-1}) \geq T$, where $\B = \text{diag}(\e_k) + \mu \LP^d$.
This is analogous to selecting a graph node with the largest set in the famous \textit{set cover} (SC) in combinatorial optimization \cite{garey99}.
The analogy between graph sampling dual \eqref{eq:samplingDual} and SC is also discussed in \cite{bai2020fast}.

To determine optimal $k$, we incrementally test each candidate $k$ from $1$, until $\lambda^-_{\min}(\S \B \S^{-1}) < T$. 
After sample node $k$ is chosen, the first $d$ nodes are grouped into a sub-graph and removed from SPG $\cG$. 
The optimization repeats for the remaining nodes until all nodes are partitioned into sub-graphs.

\subsection{Gershgorin Disc Alignment on SPG}

\begin{figure}[t]
\centering
\includegraphics[width=1.0\linewidth]{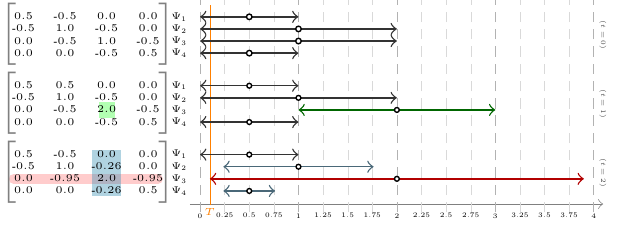}
\vspace{-0.2in}
\caption{Example of Gershgorin Disc Alignment Sampling (GDAS). 
The left shows matrix $B = \text{diag}(\a) + \mu \LP$ (for $\mu=1$) for an SPG $\cG$ with $4$ nodes and edge weights of $0.5$ between adjacent nodes.
The right shows the corresponding Gershgorin disc $\{\Psi_i\}_{i=1}^4$ for the matrix's $4$ rows.
At $t=0$, all disc left-ends $c_i - r_i$ are initially at $0$.
At $t=1$, node $3$ is sampled, and thus the corresponding disc $\Psi_3$ is shifted right by $1$. 
At $t=2$, scalar $s_3=1.9$ causes left-end of $\Psi_3$ to align at $T=0.1$, and scalar $1/s_3$ causes radii of $\Psi_2$ and $\Psi_4$ to decrease.}
\label{fig:gda}
\end{figure}

For a candidate sample node $k$, to compute an appropriate $\S = \text{diag}(\s)$ efficiently ensuring $\lambda^-_{\min}(\S (\text{diag}(\e_k) + \mu \LP^d) \S^{-1}) \geq T$ in \eqref{eq:sampleOpt} for the largest possible $d$, we perform the following recursion. 
Given sample node $k$, the $k$-th diagonal entry of $\B$ is $B_{k,k} = 1 + \mu L_{k,k}$.   
We first \textit{scale} the radius of disc $k$ of $\S \B \S^{-1}$ so that its left-end $c_k - r_k$ is \textit{aligned} at $T$; \ie, choosing scalar $s_k$ so that
\begin{align}
B_{k,k} - s_k ( |L_{k,k-1}| + |L_{k,k+1}| ) = T .
\label{eq:scalar_sk}
\end{align}
Graphically, sampling node $k$ means shifting disc $k$ right by $1$, and scaling the radius of disc $k$ with scalar $s_k$ means shifting its left-end left to $T$.
See Fig.\;\ref{fig:gda} for an illustration.

Due to $\S^{-1}$ in the similarity transform $\S \B \S^{-1}$, scaling disc $k$ means that matrix entries $L_{j,k}$ of connected nodes $j \in \{k-1,k+1\}$ will be scaled by factor $1/s_k$.
Considering first adjacent node $k+1$, 
the new radius $r_{k+1}'$ is
\begin{align}
r_{k+1}' = |L_{k+1,k+2}| + |L_{k+1,k} / s_k| .
\end{align}
If the new left-end $c_{k+1} - r'_{k+1}$ of disc $k+1$ is larger than $T$, then we next compute scalar $s_{k+1}$ so that its disc left-end is aligned at $T$, \ie, choosing $s_{k+1}$ such that
\begin{align*}
L_{k+1,k+1} - s_{k+1} \left( |L_{k+1,k+2}| + |L_{k+1,k}/s_k| \right) = T.
\end{align*}\noindent
This will in turn move the disc left-end of node $k+2$ to the right.

Eventually, the new disc left-end $c_{j} - r'_{j} < T$, meaning that the effect of sampling node $k$ has ``dissipated" to the extent that disc left-end of node $j$ can no longer move beyond $T$. 
The recursive function thus return $j-1$.

For the case when $j=k-1$, a similar disc radius scaling procedure is performed for those nodes left of $k$ progressively, until i) node $1$ is reached, in which case $0$ is returned to indicate that the left-ends of all nodes left of $k$ can move beyond $T$, or ii) the left-end of a node $j < i$ cannot move beyond $T$, in which case $-1$ is returned to signal an error. 
Algorithm~\ref{alg:LBDA} illustrates a pseudo-code of the disc alignment algorithm using recursive functions $f_L(\cdot)$ and $f_R(\cdot)$ respectively defined below

\vspace{-0.1in}
\begin{small}
\begin{align}
f_L(i,s_k) &= \left\{ \begin{array}{ll}
f_L(i-1,s_{k-1}) & \mbox{if}~~ c_i - r_i' > T, i > 1 \\
0 & \mbox{if}~~ c_i - r_i' > T, i=0 \textrm{ OR } i=0\\ 
-1 & \mbox{o.w.}
\end{array} \right.
\label{eq:f_L} \\
f_R(i,s_k) &= \left\{ \begin{array}{ll}
f_R(i+1,s_{k+1}) & \mbox{if}~~ c_{i} - r_{i}' > T \\
i-1 & \mbox{o.w.}
\end{array} \right.
\label{eq:f_R}
\end{align}
\end{small}\noindent
where new scalars $s_{k-1}$ and $s_{k+1}$ and new radii $r_i'$ are computed as described earlier. 

\begin{algorithm}[htp]
\caption{Gershgorin Disc Alignment on SPG}
\label{alg:LBDA}
\textbf{Input}: candidate sample node $k$. \\
\textbf{Output}: $d$ in \eqref{eq:sampleOpt} if feasible, $-1$ if not. \\
\vspace{-0.15in}
\begin{algorithmic}[1]
\STATE Compute scalar $s_k$ via \eqref{eq:scalar_sk}.
\STATE $L := f_L(k-1,s_k)$.
\STATE If $L = -1$, \textbf{return} $-1$.
\STATE $R := f_R(k+1,s_k)$.
\STATE \textbf{return} $R$.
\end{algorithmic}
\end{algorithm}

\vspace{-0.11in}
\section{RESULTS}
\label{sec:results}

\begin{figure*}[t]
	\centering
	\includegraphics[width=0.8\linewidth]{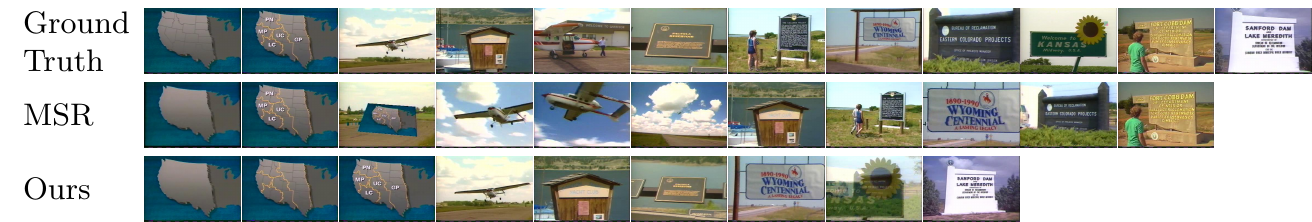}
	\caption{Selected keyframes for video ``\textit{v25.mpg}'' in VSUMM dataset in comparison to the \textit{ground-truth} (GT) and MSR~\cite{mei2015video}}\label{fig:qlt}
\end{figure*}

Evaluating video summarization techniques is challenging due to its subjective nature and limited ground-truths.
We compare our method using the VSUMM~\cite{deavila2011vsumm} benchmark dataset in quantitative and qualitative evaluations.

\subsection{Quantitative evaluation}

VSUMM \cite{deavila2011vsumm} is the primary dataset for keyframe-based video summarization. 
Collected from \textit{Open Video Project}\footnote{\url{https://open-video.org/}}~\cite{mundur2006keyframebased} (OVP), it consists of 50 videos of duration $1$-$4$ minutes and resolution of $352 \times 240$ in MPEG-1 format (mainly in 30fps).
The videos are in different genres: documentary, educational, lecture, and historical, etc.

Fifty human subjects participated in constructing user summaries, each annotating five videos. Thus, the dataset has five sets of keyframes as ground-truth per video, and each set may have a different size.
Here, we refer to computer-generated keyframes 
as \textit{automatic summary} ($\cA$), and the human-annotated keyframes as \textit{user summary} $\cU_u$, for each user $u=1,2,\dots, 5$.

For each video, the agreements between $\cA$ and each of user summaries, $\mathcal{U}_u$, are widely measured by \textit{Precision} ($P_u$), \textit{Recall} ($R_u$) and $F_{1,u}$.
{Following \cite{mei2021patcha}, we define the precision recall and $F_1$ 
against each human subject respectively as
\begin{align*}
	P_{u} = \frac{|\mathcal{A} \cap \mathcal{U}_u|}{|\mathcal{A}|},\quad 
	R_{u} = \frac{|\mathcal{A} \cap \mathcal{U}_u|}{|\mathcal{U}_u|}, \quad
	F_{1,u} = \frac{2P_u R_u}{P_u+R_u}
\end{align*}
}
where $|\mathcal{A} \cap \mathcal{U}_u|$ is the number of keyframes in $\cA$ matching  $\mathcal{U}_u$.
$|\mathcal{A}|$ and $|\mathcal{U}_u|$ are the numbers of selected keyframes in automatic and user summaries, respectively.

For each video, we take the mean precision, recall and $F_1$ across all users, $u=1,2,\dots,5$. 
The reported numbers in Table~\ref{tab:resvsumm} are the average of the above measures across all videos in the dataset.

For comparison, we follow \cite{deavila2011vsumm,mei2021patcha} and compare our method with DT~\cite{mundur2006keyframebased}, VSUMM~\cite{deavila2011vsumm}, STIMO~\cite{furini2009stimo}, MSR \cite{mei2015video}, AGDS \cite{cong2017adaptive} and SBOMP \cite{mei2021patcha}.
The results for DT, STIMO and VSUMM methods are available from the official VSUMM  website\footnote{\url{http://www.sites.google.com/site/vsummsite/}}.
Since the other algorithm implementations or their automatic summaries are not publicly available, we rely on the best results reported in \cite{mei2021patcha} for comparison.

Table\;\ref{tab:resvsumm} shows the experimental results, where the numbers in underlined bold, bold, and underlined indicate the top three performances.
SBOMP~\cite{mei2021patcha} is a recently proposed video summarization technique with state-of-the-art performance.
Our method outperformed SBOMP and performed slightly worse than its extended neighborhood mode, SBOMPn.
SBOMP relies on sub-frame representation by extracting features for each image patch collected in a \textit{matrix block}.
This leads to complex representation for video frames in their dictionary learning-based solution, increasing the algorithm's complexity---$\cO(dN^2m + d^2Nm^3)$, where $d$ is the dimension of feature vector (per patch), and $N,m$ are the number of frames and keyframes, respectively.
In contrast, our solution approaches global minimum error leveraging lemma~\ref{lemma:lemma1}, using fast recursion along a simple path graph modeling frames. 
Our complexity is $\cO(N D^2 \log_2 \frac{1}{\epsilon})$, where $D\ll N$ is the maximum depth of recursion, and $\epsilon$ is the binary search precision.
Thus, our scheme achieved similar performance at much reduced complexity.

\begin{table}\centering
\caption{Results on VSUMM}\label{tab:resvsumm}
\begin{tabular}{lllll}
		\toprule
		Algorithm & $P$ (\%) & $R$ (\%) & $F_1$ (\%) \\
		\midrule
		DT \cite{mundur2006keyframebased} & 35.51 & 26.71 & 29.43 \\
		STIMO \cite{furini2009stimo} & 34.73 & 40.03 & 35.75 \\
		VSUMM \cite{deavila2011vsumm} & \textbf{\underline{47.26}} & 42.34 & 43.52 \\
		MSR \cite{mei2015video} & 36.94 & 57.61 & 43.39 \\
AGDS \cite{cong2017adaptive} & 37.57 & \underline{64.60} & 45.52 \\
		SBOMP \cite{mei2021patcha}& 39.28 & 62.28 & \underline{46.68} \\
		SBOMPn \cite{mei2021patcha}& \textbf{41.23} & \textbf{68.47} & \textbf{\underline{49.70}} \\
Ours ($\mu=0.01,\epsilon=10^{-7}$)					& \underline{39.67} & \textbf{\underline{71.48}} & \textbf{48.92} \\\bottomrule
\end{tabular}
\end{table}

\subsection{Qualitative evaluation}

The qualitative results are evaluated based on subjective aspects such as \textit{diversity}, \textit{representativeness} and \textit{sparsity}.
Diversity indicates how dissimilar the selected keyframes are. Representativeness indicates the quality of video content coverage by a generated summary.
Keyframe based video summarization techniques strive to achieve better representativeness and diversity with the least possible number of keyframes, called \textit{sparsity}~\cite{panda2017diversityaware}.

For comparison, we follow the case study presented in \cite{mei2015video} and compare our method with their provided results in Fig.~\ref{fig:qlt}.
The ``v25'' video belongs to the documentary genre and introduces regional water projects. About a third of the video describes the regional map (the first two keyframes in the GT) and introduces the largest region (the 3rd frame), following by presenting water projects in each of its sub-divisions (the later keyframes).
Both ours and \cite{mei2015video} rely on sequentially selecting keyframes from the video.
In terms of diversity, ours has selected less redundant keyframes.
Moreover, while our algorithm has represented the video content with higher sparsity, it has better representativeness than \cite{mei2015video}.

\vspace{-0.11in}
\section{CONCLUSION}
\label{sec:conclude}
We proposed to solve the keyframe selection problem from a graph sampling perspective.  
Specifically, our method first represents the frames in a video as nodes in a similarity line graph, where edge weights are computed using features derived from pre-trained deep neural networks.
Given the constructed graph, we posed keyframe selection as a budget-constrained graph sampling problem, and devised an efficient algorithm that iteratively divides the graph into sub-graphs.
Experimental results show that our proposal achieved comparable performance to state-of-the-art methods at reduced complexity.

\bibliographystyle{IEEEbib}
% \bibliography{ref_abbriviated}

\begin{thebibliography}{10}

\bibitem{hu2011survey}
Weiming Hu, Nianhua Xie, Li~Li, Xianglin Zeng, and Stephen Maybank,
\newblock ``A {{Survey}} on {{Visual Content}}-{{Based Video Indexing}} and
  {{Retrieval}},''
\newblock {\em IEEE Trans. Syst., Man, Cybern. Syst., Part C (Applications and
  Reviews)}, vol. 41, no. 6, pp. 797--819, Nov. 2011.

\bibitem{brezeale2008automatic}
Darin Brezeale and Diane~J. Cook,
\newblock ``Automatic {{Video Classification}}: {{A Survey}} of the
  {{Literature}},''
\newblock {\em IEEE Trans. Syst., Man, Cybern. Syst., Part C (Applications and
  Reviews)}, vol. 38, no. 3, pp. 416--430, May 2008.

\bibitem{yihonggong2000video}
Yihong Gong and Xin Liu,
\newblock ``Video summarization using singular value decomposition,''
\newblock in {\em Proc. IEEE Conf. Comput. Vision Pattern Recognit.
  ({{CVPR}})}, June 2000, vol.~2, pp. 174--180 vol.2.

\bibitem{luo2009extracting}
Jiebo Luo, Christophe Papin, and Kathleen Costello,
\newblock ``Towards {{Extracting Semantically Meaningful Key Frames From
  Personal Video Clips}}: {{From Humans}} to {{Computers}},''
\newblock {\em IEEE Trans. Circuits Syst. Video Technol.}, vol. 19, no. 2, pp.
  289--301, Feb. 2009.

\bibitem{mundur2006keyframebased}
Padmavathi Mundur, Yong Rao, and Yelena Yesha,
\newblock ``Keyframe-based video summarization using {{Delaunay}} clustering,''
\newblock {\em Int. J. Digit. Lib.}, vol. 6, no. 2, pp. 219--232, Apr. 2006.

\bibitem{furini2009stimo}
Marco Furini, Filippo Geraci, Manuela Montangero, and Marco Pellegrini,
\newblock ``{{STIMO}}: {{STIll}} and {{MOving}} video storyboard for the web
  scenario,''
\newblock {\em Multimedia Tool. Appl.}, vol. 46, no. 1, pp. 47, June 2009.

\bibitem{deavila2011vsumm}
Sandra Eliza~Fontes {de Avila}, Ana Paula~Brand{\~a}o Lopes, Antonio {da Luz},
  and Arnaldo {de Albuquerque Ara{\'u}jo},
\newblock ``{{VSUMM}}: A mechanism designed to produce static video summaries
  and a novel evaluation method,''
\newblock {\em Pattern Recognit. Lett.}, vol. 32, no. 1, pp. 56--68, Jan. 2011.

\bibitem{anirudh2016diversity}
Rushil Anirudh, Ahnaf Masroor, and Pavan Turaga,
\newblock ``Diversity promoting online sampling for streaming video
  summarization,''
\newblock in {\em Proc. IEEE Int. Conf. Image Process. ({{ICIP}})}, Sept. 2016,
  pp. 3329--3333.

\bibitem{wu2017novel}
Jiaxin Wu, Sheng-hua Zhong, Jianmin Jiang, and Yunyun Yang,
\newblock ``A novel clustering method for static video summarization,''
\newblock {\em Multimedia Tool. Appl.}, vol. 76, no. 7, pp. 9625--9641, Apr.
  2017.

\bibitem{mei2021patcha}
Shaohui Mei, Mingyang Ma, Shuai Wan, Junhui Hou, Zhiyong Wang, and David~Dagan
  Feng,
\newblock ``Patch {{Based Video Summarization With Block Sparse
  Representation}},''
\newblock {\em IEEE Trans. Multimedia}, vol. 23, pp. 732--747, 2021.

\bibitem{mei2014l2}
Shaohui Mei, Genliang Guan, Zhiyong Wang, Mingyi He, Xian-Sheng Hua, and David
  Dagan~Feng,
\newblock ``L2,0 constrained sparse dictionary selection for video
  summarization,''
\newblock in {\em Proc. IEEE Int. Conf. Multimedia Expo ({{ICME}})}, July 2014,
  pp. 1--6.

\bibitem{mei2015video}
Shaohui Mei, Genliang Guan, Zhiyong Wang, Shuai Wan, Mingyi He, and David
  Dagan~Feng,
\newblock ``Video summarization via minimum sparse reconstruction,''
\newblock {\em Pattern Recognit.}, vol. 48, no. 2, pp. 522--533, Feb. 2015.

\bibitem{tillmann2015computational}
Andreas~M. Tillmann,
\newblock ``On the {{Computational Intractability}} of {{Exact}} and
  {{Approximate Dictionary Learning}},''
\newblock {\em IEEE Signal Process. Lett.}, vol. 22, no. 1, pp. 45--49, Jan.
  2015.

\bibitem{ortega18ieee}
Antonio Ortega, Pascal Frossard, Jelena Kovacevi{\'{c}}, José M.~F. Moura, and
  Pierre Vandergheynst,
\newblock ``Graph signal processing: Overview, challenges, and applications,''
\newblock {\em Proc. {IEEE}}, vol. 106, no. 5, pp. 808--828, 2018.

\bibitem{cheung18}
G.~Cheung, E.~Magli, Y.~Tanaka, and M.~Ng,
\newblock ``Graph spectral image processing,''
\newblock {\em Proc. {IEEE}}, vol. 106, no.5, pp. 907--930, May 2018.

\bibitem{bai2020fast}
Yuanchao Bai, Fen Wang, Gene Cheung, Yuji Nakatsukasa, and Wen Gao,
\newblock ``Fast graph sampling set selection using gershgorin disc
  alignment,''
\newblock {\em IEEE Trans. Signal Process.}, vol. 68, pp. 2419--2434, 2020.

\bibitem{varga04}
Richard~S. Varga,
\newblock {\em Ger$\check{\textrm{s}}$gorin and His Circles},
\newblock Springer, 2004.

\bibitem{szegedy2015going}
Christian Szegedy, Wei Liu, Yangqing Jia, Pierre Sermanet, Scott Reed, Dragomir
  Anguelov, Dumitru Erhan, Vincent Vanhoucke, and Andrew Rabinovich,
\newblock ``Going deeper with convolutions,''
\newblock in {\em Proc. IEEE Conf. Comput. Vision Pattern Recognit. ({CVPR})},
  2015, pp. 1--9.

\bibitem{deng2009imagenet}
Jia Deng, Wei Dong, Richard Socher, Li-Jia Li, Kai Li, and Li~Fei-Fei,
\newblock ``Imagenet: A large-scale hierarchical image database,''
\newblock in {\em Proc. IEEE Conf. Comput. Vision Pattern Recognit. ({CVPR})}.
  IEEE, 2009, pp. 248--255.

\bibitem{cong2017adaptive}
Yang Cong, Ji~Liu, Gan Sun, Quanzeng You, Yuncheng Li, and Jiebo Luo,
\newblock ``Adaptive greedy dictionary selection for web media summarization,''
\newblock {\em IEEE Trans. Image Process.}, vol. 26, no. 1, pp. 185--195, Jan.
  2017.

\bibitem{pang17}
Jiahao Pang and Gene Cheung,
\newblock ``Graph laplacian regularization for image denoising: Analysis in the
  continuous domain,''
\newblock {\em IEEE Trans. Image Process.}, vol. 26, no. 4, pp. 1770--1785,
  2017.

\bibitem{liu17}
X.~Liu, G.~Cheung, X.~Wu, and D.~Zhao,
\newblock ``Random walk graph laplacian based smoothness prior for soft
  decoding of {JPEG} images,''
\newblock {\em IEEE Trans. Image Process.}, vol. 26, no.2, pp. 509--524,
  February 2017.

\bibitem{chen21}
Fei Chen, Gene Cheung, and Xue Zhang,
\newblock ``Fast \& robust image interpolation using gradient graph laplacian
  regularizer,''
\newblock in {\em Proc. IEEE Int. Conf. Image Process. (ICIP)}, 2021, pp.
  1964--1968.

\bibitem{horn2012matrix}
Roger~A. Horn and Charles~R. Johnson,
\newblock {\em Matrix Analysis},
\newblock {Cambridge university press}, 2012.

\bibitem{ehrenfeld1955efficiency}
Sylvain Ehrenfeld,
\newblock ``On the efficiency of experimental designs,''
\newblock {\em Ann. Math. Stat.}, vol. 26, no. 2, pp. 247--255, 1955.

\bibitem{garey99}
M.~R. Garey and D.~S. Johnson,
\newblock {\em Computers and Intractability: A Guide to the Theory of
  {NP}-Completeness},
\newblock Freeman, 1999.

\bibitem{panda2017diversityaware}
Rameswar Panda, Niluthpol~Chowdhury Mithun, and Amit~K. {Roy-Chowdhury},
\newblock ``Diversity-aware multi-video summarization,''
\newblock {\em IEEE Trans. Image Process.}, vol. 26, no. 10, pp. 4712--4724,
  Oct. 2017.

\end{thebibliography}

\end{document}